\documentclass{SCIS}

\begin{document}
    
    \ArticleType{RESEARCH PAPER}
    \Year{2018}
    \Month{}
    \Vol{61}
    \No{}
    \DOI{}
    \ArtNo{}
    \ReceiveDate{}
    \ReviseDate{}
    \AcceptDate{}
    \OnlineDate{}
    
    \title{Event Coreference Resolution via a Multi-loss Neural Network without Using Argument Information}{Event Coreference Resolution via a Multi-loss Neural Network without Using Argument Information}
    
    \author[1,2]{Xinyu Zuo}{{xinyu.zuo@nlpr.ia.ac.cn}}
    \author[1]{Yubo Chen}{}
    \author[1,2]{Kang Liu}{}
    \author[1,2]{Jun Zhao}{}
    
    \AuthorMark{Xinyu Zuo}
    
    \AuthorCitation{Xinyu Zuo, Yubo Chen, Kang Liu, et al}
    
    
    \address[1]{Institute of Automation, Chinese Academy of Sciences, Beijing {\rm 100190}, China}
    \address[2]{University of Chinese Academy of Sciences, Beijing {\rm 100049}, China}
    
    \abstract{Event coreference resolution(ECR) is an important task in Natural Language Processing (NLP) and nearly all the existing approaches to this task  rely on event argument information. However, these methods tend to suffer from error propagation from the stage of event argument extraction. Besides, not every event mention contains all arguments of an event and argument information may confuse the model that events have arguments to detect event coreference in real text. Furthermore, the context information of an event is useful to infer coreference between events. Thus, in order to reduce the errors propagated from event argument extraction and use context information effectively, we propose a multi-loss neural network model which does not need any argument information to do the within-document event coreference resolution task and achieve a significant performance than the state-of-the-art methods.}
    
    
    \maketitle

    \section{Introduction}
    Event coreference resolution (ECR) is a task about determining which event mentions in a document refer to the same real-world event. Event coreference resolution is an important part of NLP systems such as summarization~\cite{daniel2003sub}, text-level event extraction~\cite{humphreys1997event}, question answering~\cite{narayanan2004question} and so on. Besides, compared to considerable research of entity coreference resolution, there is less attention on event coreference resolution. Therefore, event coreference resolution is still a challenging task and the performance should be improved. 
    
\begin{figure}[!t]
        \includegraphics[width=\textwidth]{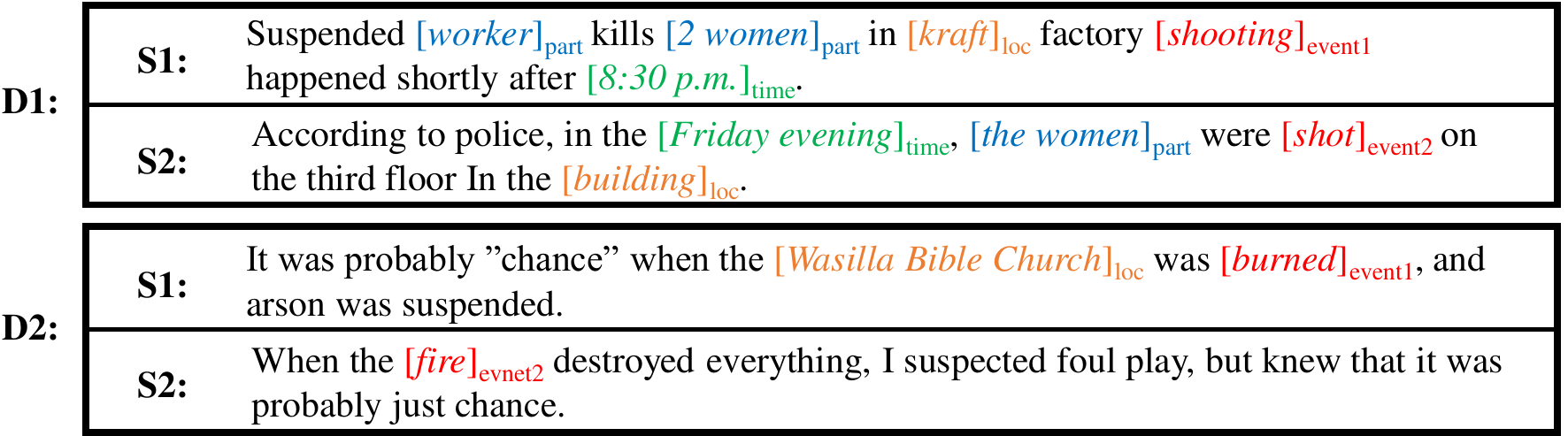}
        \caption{Event coreference resolution instances.} \label{fig1}
    \end{figure}

    Event mentions that refer to the same event can occur both within a document (WD) and across multiple documents (CD). We focus on WD event coreference in this paper because WD event coreference is the basic work of CD event coreference.
    The main task of WD event coreference is judging whether a pair of events are coreferential or not.
    Figure~\ref{fig1} shows two coreferential event pairs from two documents. The first event pair in \textbf{D1} is about \emph{shooting} event and the second event pair in \textbf{D2} is about \emph{fire} event.

    In order to judge the coreference of a event pair, most approaches for solving event coreference resolution relied on various linguistic properties especially \emph{event argument}, which contains {spatio-temporal} information of events\cite{bejan2010unsupervised}. For instances, in Figure~\ref{fig1}, the words with red front are events. And the words with blue, green and orange front are participant, time, location of the events respectively.

    Although event arguments contain useful information for event coreference resolution, there are two problems for using event arguments information in event coreference resolution. Firstly, it's difficult to extract event arguments accurately due to the diversity of the expression of event arguments. The performance of event argument extraction is only 55.7$\%$~\cite{chen2017automatically} in ACE corpus. For instance, in \textbf{D1}, the arguments about \emph{shooting} event in the two sentences are the same but are expressed differently. In details, in \textbf{D1}, the participant, time, location of \emph{shooting} event are \emph{worker/2 women}, \emph{8:30 p.m.} and \emph{kraft} in \textbf{S1}, but \emph{the women}, \emph{Friday evening} and \emph{building} in \textbf{S2} respectively. Secondly, not every event mention contains all arguments of one event that may make model confused about the coreference of two events in a event pair. For instance, in \textbf{D2}, the \emph{Wasilla Bible Church} that the location of \emph{fire} event is in \textbf{S1} but not in \textbf{S2}. Besides, in \textbf{D2}, devoid of event arguments, \emph{burned} event and \emph{fire} event are coreferential in context.

    As aforementioned, the arguments of events are difficult to extract. It is also difficult to use arguments to solve all the problems of event coreference resolution even if they are extracted. Thus, the context about event mentions is more important and effective for event coreference resolution. In order to use context information efficiently, we propose a multi-loss neural network model (MLNN) which doesn't need any argument information to accomplish within-document event coreference resolution task. We propose two sub-models which use context information to detect the coreference of two events in a event pair and train them jointly. One is a classifier which  predicts whether the two events in one pair are coreferential, and another is a scorer which calculates similarity scores between them to assist infer coreference. 
	
    The final stage about event coreference resolution is event clustering. After all event pairs are predicted and scored, we filter event pairs according to the results of classifier and scorer. Then, we use a dynamic connectivity algorithm to construct a graph for event clustering. Each node in graph is a event mention and each edge between two nodes represent whether the two event are coreferential or not. Finally, all events connected in one graph are considered to be in one event cluster(event chain).
	
    We evaluate our model on ECB+ corpus~\cite{cybulska2014guidelines} and use \emph{B$^3$}~\cite{bagga1998algorithms}, \emph{CEAF$_e$}~\cite{luo2005coreference}, \emph{MUC}~\cite{vilain1995model} and \emph{CoNLL $F_1$}~\cite{pradhan2014scoring} as measures. The experimental results show that our model achieve a significant improvement compared to the state-of-the-art methods which use event argument features. 

    \section{Task Description}
    We adopt ECB+ corpus which extends the widely used corpus for event coreference resolution task,  EventCorefBank (ECB)~\cite{bejan2010unsupervised}. An \textbf{event} is something happens or a situation occurs in real world~\cite{cybulska2014using}. In ECB+ corpus, a event consists of four components: (1) an \emph{Action}: what happens in the event; (2) \emph{Participants}: who or what is involved; (3) a \emph{Time}: when the event happens; and (4) a \emph{Location}: where the event happens. Each document consists of a set of mentions which describe event actions, participants, times, and locations. These mentions relate to different events in a document. Table~\ref{tab1} shows instances about \emph{shooting} event components of the sentence "\emph{Suspended worker kills 2 women in kraft factory shooting happened shortly after 8:30 p.m..}" in \textbf{D1} that shown in Figure~\ref{fig1}. 
    
    \begin{table}[!t]
        \centering
        \caption{Mentions of event components in ECB+ corpus}
        \label{tab1}
        \tabcolsep 33pt
        \begin{tabular*}{\textwidth}{cccc}
        \toprule
        Action  & Participant & Time & Location \\\hline
        shooting & worker/2 women & 8:30 p.m. & kraft\\
        \bottomrule
        \end{tabular*}
    \end{table}

    In order to consist with the ECB+ corpus, in our paper, we also use the term \textbf{event mention} which is usually verb or noun phrase that describe events most clearly to refer to the mention of an event \emph{action}, and \textbf{event argument} to refer to all mentions of the \emph{participant}, \emph{time}, and \emph{location} included in the event. Additionally, we define that two events is coreferential when they refer to the same actual event in real world. As we can see in Figure~\ref{fig1}, although we can infer coreference easily if we use all event arguments information of an event, not all event arguments are presented in an event. Thus, we need to reduce the errors propagated from this issue and event argument extraction, and need to utilize the context about events, which is the most reliable information for infering coreference.

    \section{Related work}
    Coreference resolution in general is a difficult natural language processing(NLP) task and typically requires sophisticated inferentially-based knowledge-intensive models~\cite{kehler2002coherence}. Extensive work in the literature focuses on the problem of entity coreference resolution and many techniques have been developed, including rule-based deterministic models~\cite{cardie1999noun} which traverse over mentions in certain orderings and make deterministic coreference decisions based on all available information at the time; supervised learning-based models~\cite{stoyanov2009conundrums} which make use of rich linguistic features and the annotated corpora to learn more powerful functions.

    Event coreference resolution is a more complex task than entity coreference resolution~\cite{humphreys1997event} and also has been relatively less studied. Different approaches have been proposed to detect within-document coreference chains. Works specific to within-document event coreference include pairwise classifiers~\cite{ahn2006stages,chen2009pairwise,zeng2018adversarial} graph based clustering method~\cite{chen2009graph}, information propagation~\cite{liu2014supervised}, markov logic network~\cite{lu2016joint}, linguistic features based on unsupervised method~\cite{bejan2010unsupervised}, hierarchical distance-dependent bayesian model~\cite{yang2015hierarchical} and iterative unfolding inter-dependencies model~\cite{choubey2017event}.

    Like these works, almost all well-performed methods rely on rich features. These methods require complex and time-consuming feature engineering and bring more propagation of errors.

    \section{Methodology}
    The event coreference resolution task in this paper can divided into a main sub-task and two secondary sub-tasks: (1) (main sub-task) \textbf{event coreference detection}: detecting whether each candidate event pair is coreferential or not, (2) (secondary sub-task) \textbf{event mention extraction}: extracting event mentions, and (3) (secondary sub-task) \textbf{event clustering}: grouping event mentions into clusters according to the coreference of them.

    \subsection{Event Mention Extraction Method}
    Previous methods about event coreference resolution rely on rich features based on semi-Markov CRFs~\cite{yang2015hierarchical} to identify event mentions. The features include word-level features, such as unigrams, bigrams, POS tags, WordNet hypernyms, synonyms and FrameNet semantic roles, and phrase-level features such as phrasal syntax(e.g. NP, VP) and phrasal embeddings (constructed by averaging word embeddings). Based on head word matching\footnote{For multi-word event mentions, in order to be same with previous methods, we only use the first word and its word embedding to represent event mentions.}, 95$\%$ event mentions can be identified in development set.

    In order to consist with the event coreference detection model and use less features, we build a multi-layer feed-forward neural network with cross-entropy objective function to identify whether a candidate word is an event mention or not. Additionally, our neural network only uses candidate word, context in a window around candidate word, POS tags in a window around candidate word and the lemma of candidate word as features. Our model can identify 92$\%$ event mentions in the same development set as Yang et al. (2015), slightly lower than semi-Markov CRFs.

    \begin{figure}[!t]
        \includegraphics[width=\textwidth]{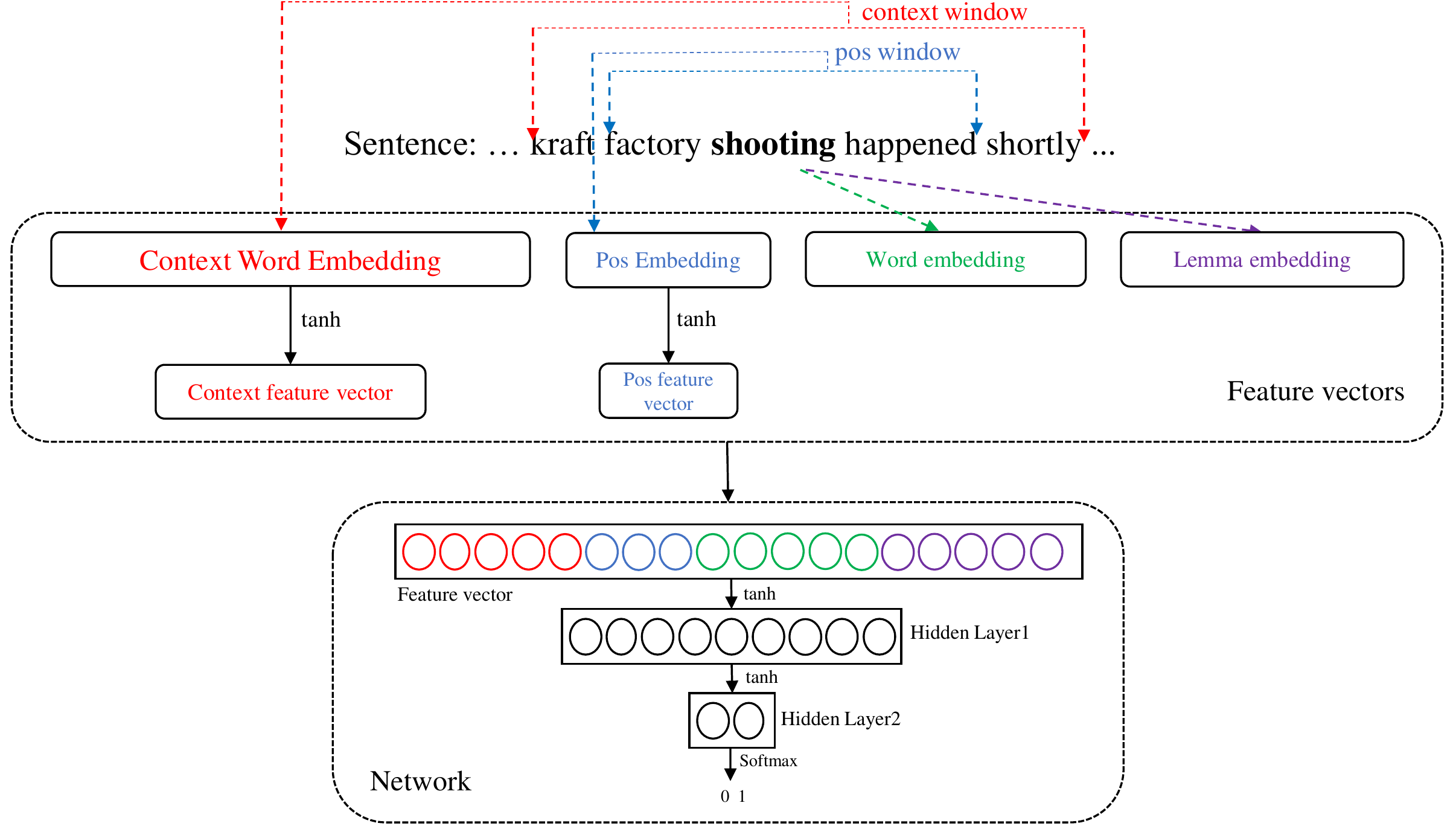}
        \caption{Feed-forward Neural Network structure for Event Mention Extraction} \label{fig2}
    \end{figure}

    As Figure~\ref{fig2} shows, we treat the event mention extraction task as a classification task. For each candidate word, \textbf{firstly}, we take the features aforementioned as input and convert them into context embedding (a combination of words in the window), POS embedding (a combination of POS tags in the window), word embedding, lemma embedding respectively. \textbf{Secondly}, we map context word embedding and POS embedding into a context feature vector and a POS feature vector respectively by one layer feed-forward NN. \textbf{Thirdly}, we combine all feature embeddings and pass them through the two layers feed-forward NN which uses tanh as activation function. \textbf{Finally}, the model output a two dimensions vector that the value of each dimension is 0 or 1 after softmax operation. The candidate word will be predicted as an event mention if the value of first dimension is 1, and will be predicted as a non-event mention if the value of second dimension is 1.

    \subsection{Event Coreference Detection Method}
    We construct a multi-loss neural network(MLNN) model which needs no event arguments information and trains classifier and scorer jointly. The input of the network are a candidate event pair and its features. And, the system outputs a classification result which indicates whether a event pair is coreferential or not preliminary, a confidence score and similarity score assist us to infer coreference eventually.

    \begin{figure}[!t]
	    \includegraphics[width=\textwidth]{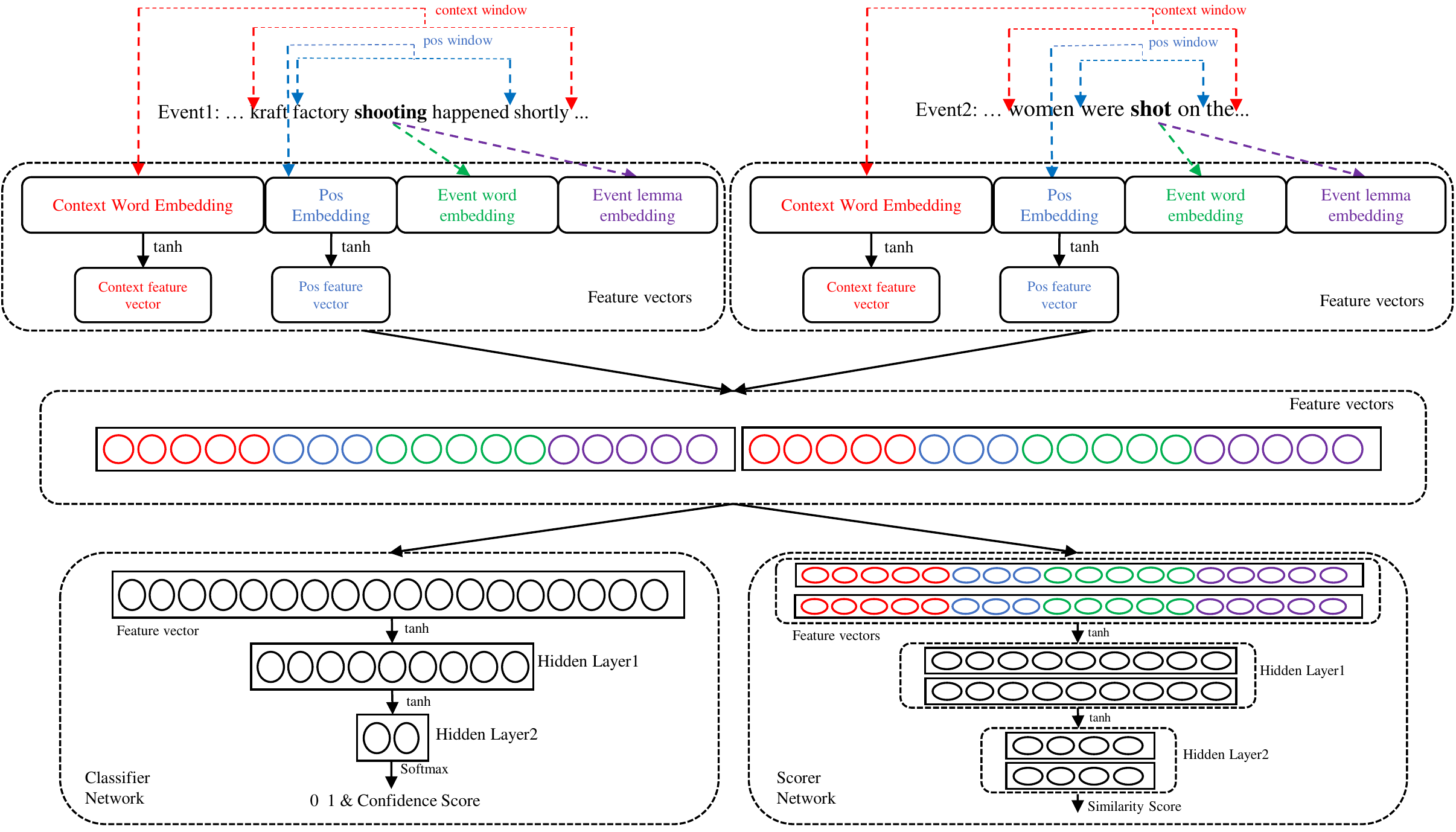}
	    \caption{Multi-loss Neural Network(MLNN) structure for Event Coreference Detection} \label{fig3}
    \end{figure}

    \subsubsection{Event Features}
    We use similar features as the event mention extraction: (1) \emph{context feature}: the context around the candidate event in a window, (2) \emph{POS feature}: the POS tags of words around the candidate event in a window, and (3) \emph{lexical feature}: the word and its lemma of the candidate event. Same as event mention extraction, we convert context feature and POS feature into context word embedding and POS embedding respectively, then map them into a context feature vector and a POS feature vector respectively by one layer feed-forward NN. Additionally, we convert lexical feature into a {event word embedding and a event lemma embedding. Finally, we combine context feature vector, POS feature vector, event word embedding and event lemma embedding as a feature vector about the candidate event.

    \subsubsection{Multi-loss Neural Network}
    Figure~\ref{fig3} shows the structure of multi-loss neural network (MLNN) for event coreference detection.

    \emph{\textbf{Classification Model with Cross-entropy Objective Function}}: The first sub-network is \emph{classifier network(CN)} in Figure~\ref{fig1}. \textbf{Firstly}, We combine the feature vectors of two events in a candidate event pair as one feature vector and input into CN. \textbf{Secondly}, we pass the combined feature vector through two layers feed-forward NN which uses tanh as activation function. \textbf{Finally}, after a softmax operation we get a result of classification which indicates whether the two events in a candidate event pair is coreferential or not and a confidence score assists us to infer coreference. Moreover, the cross-entropy objective function of CN is as following:

    \begin{equation}
        \mathcal{L}_1(\theta_1) = -\frac{1}{n}\sum_{i=0}^n[y_i\log \widehat{y_i} + (1-y_i)\log (1-\widehat{y_i})] \quad  y_i, \widehat{y_i} \in \left\{0, 1 \right\}
    \end{equation}
    
    \begin{equation}
        \widehat{y_i} = P(y_i = 1|\textbf{x}_i), 1-\widehat{y_i} = P(y_i = 0|\textbf{x}_i)
    \end{equation}
    
    Where $\textbf{x}_i$ is a input candidate event pair, $y_i$ and $\widehat{y_i}$ are the correct and predicted label which indicate the coreference of two event in a candidate event pair respectively. It indicates a coreferential event pair if the value is 0 and not coreferential if the value is 1. Additionally, $n$ is the quantity of input event pairs and $\theta_1$ is the parameter of CN.
    
    \emph{\textbf{Scoring Model with Similarity Difference Objective Function}}: The second sub-network is \emph{scorer network (SN)} in figure 3 which is different from CN. \textbf{Firstly}, we input the feature vector of two events in a candidate event pair to SN individually rather than combine them. \textbf{Secondly}, we pass the feature vectors of two events through a two layers feed-forward NN which uses tanh as activation function. \textbf{Finally}, we calculate the cosine similarity score of two vectors which is the output of two layers NN.  Moreover, the similarity difference objective function of SN is following:
    
    \begin{equation}
        \mathcal{L}_2(\theta_2) = \sum_{i=0}^n\log |m_i - s_i|, 
        \left\{  
        \begin{array}{lr}  
            m_i=1, \quad if \quad y_i = 0 &  \\  
            m_i=-1, \quad if \quad y_i = 1 &    
        \end{array}  
        \right.  
     \end{equation}  
     
     Where $s_i$ is the cosine similarity score of two events in the input event pair. The $s_i$ is closer to 1, the more similar the two events are, and is closer to -1, the less similar the two events are. And $m_i$ is the margin of cosine similarity score.
    
    \emph{\textbf{Jointly Training with Multi-loss Function}}: We train the two sub-model jointly and combine their objective function as following:
    
     \begin{equation}
         \mathcal{L}_{all}(\theta_{all}) = \mathcal{L}_1(\theta_1) + \mathcal{L}_2(\theta_2)
     \end{equation}
     
    Where the $\theta_{all}$ are parameters of the whole system.

    \subsection{Event Clustering Method}

    After classification and scoring, we filter event pairs according to the result obtained from classifier and scorer. After that, we construct a graph to cluster the filtered events by a dynamic connectivity algorithm. Each node in graph is a event mention and the two events is coreferential if there is a edge between them. The details as the following:

    \subsubsection{Event Pair Filtering} The annotations in ECB+ about event mention and event coreference is incomplete because some coreferences between events in real world are not marked in text.~\cite{cybulska2014guidelines}. 
    This phenomenon propagates an error that a word annotated as event mention but its not in fact. Thus, we use confidence and similarity scores to enhance recall about event coreference detection.
    Eventually, for each input candidate event pair, the two events are coreferential if the classification result is \emph{coreference}. We also identify the two events are coreferential if the classification result is \emph{non-coreference} but similarity score is greater than 0.5 and confidence score is less than 0.6. The threshold of similarity score and confidence score are determined by development set.

    \subsubsection{Event Clustering} We use dynamic connectivity algorithm to merge two events which we identify coreferential in each event pair from one document. After merging, we regard each individual subgraph as a cluster, and events in a same cluster as a coreferential event chain. 

    \section{Evaluation}
    We perform all the experiments on the ECB+ corpus. We adopt the settings about datasets used in Choubey et al. (2017). We divide the dataset into training set (topics 1-20), development set (topics 21-23) and test set (topics 34-43) which are same as Choubey et al. (2017). Table~\ref{tab2} shows the distribution of corpus.

    \begin{table}[!t]
    \caption{ECB+ corpus statistics}
    \label{tab2}
    \tabcolsep 25pt 
    \begin{tabular*}{\textwidth}{ccccc}
    \toprule
       & Train & Dev & Test & Total \\\hline
       \#Document & 462 & 73 & 447 & 982\\
       \#Sentences & 7294 & 649 & 7867 & 15810\\
       \#Event Mentions & 3555 & 441 & 3290 & 7268\\
       \#WD Chains & 2499 & 316 & 2137 & 4953\\
       Avg. WD chain length & 2.8 & 2.6 & 2.6 & 2.7\\
    \bottomrule
    \end{tabular*}
    \end{table}

    We evaluate our system with four widely used coreference resolution metrics: \emph{B$^3$} measures the proportion of overlap between the predicted and gold clusters for each mention, \emph{CEAF$_e$} measures the best alignment of the gold-standard and predicted clusters, \emph{MUC} measures that how many gold cluster merging operations are needed to recover each predicted cluster and \emph{CoNLL $F_1$}, the most important metric, is the average of the \emph{$F_1$} scores about all the three metrics.

    We use Natural Language ToolKit (NLTK)~\cite{bird2009natural} tools to extract POS and lemma. In details, we set the window size of context as 5 and the window size of POS as 3. Moreover, we set the size of word embedding, POS embedding, lemma embedding as 100, 10, 100 respectively. Besides, we minimize the objective function over shuffled mini-batches with the Adadelta~\cite{zeiler2012adadelta} update rule and use the publicly available official implementation of revised coreference scorer (v8.0.1)\footnote{https://github.com/conll/reference-coreference-scorers}.

    \subsection{Baseline and our Systems}
    We compare our model with five baselines\footnote{The results are taken from the Coubey et al. (2017)}.

    LEMMA: The first baseline groups event mentions into clusters if they have the same lemmatized head word. This is considered as a strong baseline. 

    HDDCRP~\cite{yang2015hierarchical}: The second baseline is the supervised Hierarchical Distance Dependent Bayesian Model on the ECB+ corpus. This model utilizes distances between event mentions, generated using a feature-rich learnable distance function, as Bayesian priors for single pass non-parametric clustering.

    HDP-LEX~\cite{bejan2010unsupervised}: The third baseline is a unsupervised hierarchical bayesian model by Bejan (2010)

    Agglomerative~\cite{chen2009pairwise}: The fourth baseline is a two step agglomerative clustering model.

    Iterative WD/CD Classifier(Iter-WD/CD)~\cite{choubey2017event}: This baseline is a iterative event coreference model by exploiting inter-dependencies with both WD and CD event mentions.

    All five baselines use event argument features, and the five models are tested on the same data set as above.

    \textbf{Our system}: Firstly, we detect and filter the event mentions annotated in corpus by our event mention extraction model before event coreference detection and clustering, because the annotations about event mentions and coreference is incomplete. Secondly, in order to show the performance about our multi-loss neural network, we evaluate three different systems about event coreference detection and clustering. (1) \textbf{C-NN}: We only train classification model with cross-entropy objective function and use the results of classifier to infer coreference. (2) \textbf{C-MLNN}: We train the full MLNN model, but only use the results of classifier to infer coreference. (3) \textbf{MLNN}: We use full MLNN model with the results of classifier and scorer (\emph{coreference} classification result and \emph{non-coreference} classification result with confidence and similarity scores).

    \subsection{Results}

    \begin{table}[]
        \centering
        \caption{Within-document event coreference resolution results on ECB+ corpus}
        \label{tab3}
        \tabcolsep 6pt
        \begin{tabular*}{\textwidth}{c|ccc|ccc|ccc|c}
            \toprule
            & \multicolumn{3}{c|}{$B^3$}                          & \multicolumn{3}{c|}{MUC}                       & \multicolumn{3}{c|}{$CEAFE_e$}                    & CoNLL $F_1$         \\ \hline
            & R              & P             & $F_1$             & R              & P             & $F_1$            & R             & P             & $F_1$            & $F_1$            \\ \hline
            LEMMA         & 56.8           & \textit{80.9} & 66.7           & 35.9           & \textbf{76.2} & 48.8          & 67.4          & 62.9          & 65.1          & 60.2          \\ 
            HDP-LEX (2010)       & 67.6           & 74.7          & 71.0           & 39.1           & 50.0          & 43.9          & 71.4 & 66.2          & 68.7          & 61.2          \\ 
            Agglomerative (2009) & 67.6           & 80.7          & 73.5           & 39.2           & 61.9          & 48.0          & \textit{76.0} & 65.6          & 70.4          & 63.9          \\ 
            HDDCRP (2015)        & 67.3           & \textbf{85.6} & \textit{75.4}  & 41.7           & \textit{74.3} & 53.4          & \textbf{79.8} & 65.1          & \textit{71.7} & 66.8          \\ 
            Iter-WD/CD (2017)    & \textit{69.2}  & 76.0          & 72.4           & \textit{58.5}  & 67.3          & \textbf{62.6} & 67.9          & \textbf{76.1} & \textbf{71.8} & \textit{68.9} \\ 
            \textbf{MLNN}   & \textbf{87.3} & 71.0         & \textbf{78.3} & \textbf{69.0} & 57.0         & \textit{62.4} & 66.6          & \textit{76.0} & 70.7         & \textbf{70.4} \\ 
            \bottomrule
        \end{tabular*}
    \end{table}
    
    Table~\ref{tab3} shows the results about within document event coreference resolution on ECB+ corpus. In Table~\ref{tab3}, the bold results are the best performance of each metrics, and the italic results are the second best performance of each metrics. We notice that, although we do not use any features about event argument which is a very important information when understanding events, and the CRFs method which other systems used performs a little better than us in event mention extraction, the performance of \emph{CoNLL $F_1$} about our \emph{MLNN} model is obviously better than the state-of-the-art methods. In details, although the $F_1$ value in \emph{CEAF$_e$} is slightly lower, we obtain the second best $F_1$ value in \emph{MUC} and the best $F_1$ value in \emph{B$^3$}. On the whole, there is a obvious improvement of the performance of within-document event coreference resolution with MLNN model.
    
    \begin{table}[]
        \centering
        \caption{Comparisons about three systems}
        \label{tab4}
        \tabcolsep 8pt
        \begin{tabular*}{\textwidth}{c|ccc|ccc|ccc|c}
            \toprule
            & \multicolumn{3}{c|}{$B^3$}                        & \multicolumn{3}{c|}{MUC}                      & \multicolumn{3}{c|}{$CEAFE_e$}                    & CoNLL $F_1$         \\ \hline
            & R             & P             & $F_1$            & R             & P             & $F_1$            & R             & P             & $F_1$            & $F_1$            \\ \hline
            C-NN   & \textbf{90.2} & 48.8          & 63.3          & \textbf{76.8} & 40.0          & 56.0          & 40.2          & 69.7          & 51.0          & 56.8          \\ 
            C-MLNN & 86.8          & 67.7          & 76.0          & 67.6          & 53.3          & 59.6          & 62.3          & 74.5          & 67.9          & 67.8          \\ 
            MLNN   & 87.3          & \textbf{71.0} & \textbf{78.3} & 69.0          & \textbf{57.0} & \textbf{62.4} & \textbf{66.6} & \textbf{76.0} & \textbf{70.7} & \textbf{70.4} \\ 
            \bottomrule
        \end{tabular*}
    \end{table}

    In Table~\ref{tab4}, \textbf{firstly}, the results of \emph{C-NN} and \emph{C-MLNN} indicate that classification model training with scoring model can improve the performance of classifier to infer event coreference obviously. \textbf{Secondly}, the results of \emph{C-MLNN} and \emph{MLNN} show that the scores from scorer can assist classifier to improve the performance of within-document event coreference resolution. 

    \section{Errors Analysis}

    There are two main issues cause errors.

    \subsection{Wrong coreference links} As our method utilizes contextual information mainly, events in adjacent positions in the same sentence which from different event chains may be inferred as coreferential events. For instane, in sentence \emph{"This anti-piracy action by INS Sukanya was the fifth successful operation by its crew during its current \textbf{patrol} \textbf{mission} in the Gulf of Aden since September this year."}, the two events, \emph{patrol} event and \emph{mission} event are not coreferential but have similar context.

    \subsection{Missed annotation} As mentioned above, the annotations of event mention and coreference is incomplete, for example, in sentence \emph{"INS Sukanya has \textbf{sized} a total of 14 AK-47 rifles, 31 magazines and 923 rounds of ammunition during the five operations it carried out there, the officer said."}, \emph{sized} is a event mention but not marked in ECB+ corpus.

    \section{Conclusion and Future Works}
    We present a multi-layer feed-forward neural network for event mention extraction and a multi-loss neural network model for within-document event coreference resolution respectively. We do not use any information about event argument in our system. Additionally, we test our system in ECB+ corpus and achieve a significant improvement than the state-of-the-art methods.

    Due to the incomplete annotation and the propagation of errors about event mentions and arguments extraction in pipeline systems, we will try to design a joint model to accomplish the event extraction, argument extraction, event coreference resolution tasks jointly in the future.

    \Acknowledgements{This work is supported by the Natural Science Foundation of China (No.61533018, No.61702512 and No.61502493). This work was also supported by Alibaba Group through Alibaba Innovative Research (AIR) Program and Huawei Tech. Ltm through Huawei Innovation Research Program.).}

    
    \bibliographystyle{splncs04}
    \bibliography{mybibliography}
    
    
    
    
    
    
    \end{document}